%% file: main.tex
\documentclass[10pt, journal]{IEEEtran}
\IEEEoverridecommandlockouts

\usepackage{cite}

\ifCLASSINFOpdf
  \usepackage[pdftex]{graphicx}
  \graphicspath{{../pdf/}{../jpeg/}}
  \DeclareGraphicsExtensions{.pdf,.jpeg,.png}
\else
  \usepackage[dvips]{graphicx}
  \graphicspath{{../eps/}}
  \DeclareGraphicsExtensions{.eps}
\fi
\usepackage{amsmath}
\usepackage{array}

\ifCLASSOPTIONcompsoc
  \usepackage[caption=false,font=normalsize,labelfont=sf,textfont=sf]{subfig}
\else
  \usepackage[caption=false,font=footnotesize]{subfig}
\fi
\usepackage{microtype}
\usepackage{soul}
\usepackage{booktabs} 
\usepackage{siunitx}
\usepackage{csquotes}
\usepackage{multirow}
\usepackage{pgfplotstable}
\usetikzlibrary{patterns}
\pgfplotsset{compat=newest}
\usepackage{url}
\usepackage{hyperref}
\def\BibTeX{{\rm B\kern-.05em{\sc i\kern-.025em b}\kern-.08em
		T\kern-.1667em\lower.7ex\hbox{E}\kern-.125emX}}
		
\begin{document}
	\title{Cluster-based Input Weight Initialization\\for Echo State Networks}
	
	\author{Peter~Steiner,~\IEEEmembership{}
		Azarakhsh~Jalalvand~\IEEEmembership{Member,~IEEE,}
		Peter~Birkholz,~\IEEEmembership{Member,~IEEE}
		\thanks{This research was financed by Europ\"aischer Sozialfonds (ESF), the Free State of Saxony (Application number: 100327771), and Special Research Fund of Ghent University (BOF19/PDO/134). \textit{(Corresponding author: Peter Steiner)}}%
		\thanks{P. Steiner and P. Birkholz are with the Institute for Acoustics and Speech Communication, Technische Universität Dresden, 01069 Dresden, Germany, (e-mail: peter.steiner@tu-dresden.de; peter.birkholz@tu-dresden.de)}
		\thanks{Azarakhsh Jalalvand is with IDLab, Ghent University, Belgium, as well as Mechanical and Aerospace Engineering department, Princeton University, USA (email: azarakhsh.jalalvand@ugent.be)}
		\thanks{Manuscript received MONTH XX, 2021; revised MONTH XX, 2021.}
	}

	\markboth{IEEE TRANSACTIONS ON NEURAL NETWORKS AND LEARNING SYSTEMS,~Vol.~XX, No.~XX, MONTH\_XX~2022}%
	{Steiner \MakeLowercase{\textit{et al.}}: Cluster-based Input Weight Initialization for Echo State Networks}
	%

	\maketitle
	
	\begin{abstract}
	\boldmath
	Echo State Networks (ESNs) are a special type of recurrent neural networks (RNNs), in which the input and recurrent connections are traditionally generated randomly, and only the output weights are trained. Despite the recent success of ESNs in various tasks of audio, image and radar recognition, we postulate that a purely random initialization is not the ideal way of initializing ESNs. The aim of this work is to propose an unsupervised initialization of the input connections using the $K$-Means algorithm on the training data. We show that for a large variety of datasets this initialization performs equivalently or superior than a randomly initialized ESN whilst needing significantly less reservoir neurons. Furthermore, we discuss that this approach provides the opportunity to estimate a suitable size of the reservoir based on prior knowledge about the data. 
	\end{abstract}
	
	
	\section{Introduction}
	\label{sec:Introduction}
	
	Since the breakthrough of ESNs~\cite{src:Jaeger-01a}, a lot of design strategies for ESNs have been proposed. Although randomly initialized ESNs have achieved state-of-the-art results in various directions, several publications, such as~\cite{src:Ozturk-07, src:Lukosevicius-12b, src:Scardapane-17} argue that there should exist better approaches that incorporate more prior or biologically plausible knowledge. According to~\cite{src:Xue-07}, it requires a lot of trial and error to initialize an ESN for a task, and the relationship between the different weight matrices is not completely understood.
	
	To better understand the behavior of randomly initialized ESNs for digit and phoneme recognition, Jalalvand et. al.\ \cite{src:Jalalvand-15b, src:Jalalvand-18, src:Bala-18} have analyzed an ESN with optimized hyper-parameters and determined the impact of the different hyper-parameters. For example, it turned out that even very sparse weight matrices are still sufficient for achieving proper results. Similarly, in \cite{src:Aceituno:20}, it was shown that the hyper-parameters should be tuned to match the spectral properties of the reservoir states and the target outputs. Another way to design ESNs more efficiently is to concatenate multiple reservoirs. Different architectures, such as layered ESNs \cite{src:Jalalvand-15b} or deep and tree ESNs \cite{src:Gallicchio-17, src:Gallicchio-18, src:Gallicchio-20, src:Pathak-18} were shown to improve the performance over a single-layer ESN whilst reducing the training complexity by utilizing smaller reservoirs in each layer.
	
	Other approaches step even farther away from random initialization. The publications~\cite{src:Rodan-11, src:Wolf-12,src:Griffith-19} proposed \textit{simple} ESN reservoirs in different flavors, e.g.\ delay lines with optional feedback, cyclic reservoirs, or even a simple chain of neurons. It was shown that the proposed reservoir design strategies outperformed randomly initialized ESNs in various aspects, such as classification or regression accuracy and in terms of memory capacity. A big advantage is a relatively high sparsity, which is memory-efficient and computationally cheap. However, at least in~\cite{src:Wolf-12}, the authors warned that their findings might not hold in practice, when one needs to deal with high-dimensional inputs or more complex tasks. A related approach to simplify the reservoir initialization was proposed in \cite{src:Carrol-19}, where the reservoir weights are only allowed to have the values \num{0} and $\pm\num{1}$. Starting with the values \num{1} and \num{0}, which were assigned to the reservoir in a deterministic way, several \num{1} weights were flipped to \num{-1} and the authors have shown that this strongly influenced the behaviour of the reservoir in terms of fitting error and memory capacity. In~\cite{src:Appeltant-11}, the aforementioned approaches were further simplified, and only one neuron was used in the reservoir. The output of the neuron was fed back to its input using different delay times. This produced virtual nodes that simulated a larger reservoir.
	
	All these pioneering approaches try to initialize the reservoir and/or input weights in a more or less deterministic way that is almost task-independent. Alternative techniques also aim to initialize the ESN in a deterministic way that is, however, more task-dependent or dependent on the input data. For example,~\cite{src:Lukosevicius-10} adopted recurrent self-organizing maps (SOMs) to initialize the input and recurrent weight matrices using the SOM algorithm. Therefore, a new neuron model was used, and the weight matrices were pre-trained using the unsupervised SOM algorithm. In~\cite{src:Basterrech-11} scale-invariant maps (SIMs), an extension of SOMs, were used to initialize the input weights. The same group also used Hebbian learning in~\cite{src:Basterrech-13}. Lazar~et. al.~\cite{src:Lazar-09} proposed a biologically-inspired self-organizing recurrent neural network (SORN) consisting of spiking neuron models, in which the weights of frequently firing neurons are increased during training. This was adopted for the Batch Intrinsic Plasticity (BIP) for ESNs~\cite{src:Schrauwen-08}, where the reservoir weights were iteratively pre-trained.
	
	Yet another family of unsupervised learning algorithms are clustering techniques. In~\cite{src:Ashour-16}, the Inverse Weighted $K$-Means (IWK) algorithm~\cite{src:Barbakh-08a, src:Barbakh-08b} was proposed to initialize the weight matrices of an ESN. After randomly initializing the input weights, they applied IWK to the neuron inputs and adapted the input weights. Then they randomly initialized the recurrent weights and applied IWK again on the reservoir states to adapt the recurrent weights. The authors showed that their method outperformed a randomly initialized ESN and that the performance gets more stable when repeating random initialization. 
	
	In this paper, we present an alternative strategy to initialize the ESN input weights using the $K$-Means algorithm. Instead of applying the cluster algorithm on the neuron inputs, we used the $K$-Means algorithm to cluster the \textit{input features} and used the centroid vectors as input weights.  
	
	The main advantages of our approach are:
	\begin{itemize}
    	\item The pre-trained ESNs perform equally well or better than ESNs without pre-training and need smaller reservoirs. This will be discussed in detail on a large-scale video dataset and evaluated on a broad variety of datasets with different characteristics.
    	\item We show that the same hyper-parameter optimization strategy proposed in ~\cite{src:Jalalvand-15b,src:Jalalvand-18} for conventional ESNs can be applied to optimize the hyper-parameters of the novel $K$-Means-based initialized ESN (KM-ESN). 
	    \item Applying clustering techniques is a common data exploration step to study possible correlations within the data. Our approach efficiently benefits from the outcome of this step to also initialize the input weights.
	    \item Since the clusters are usually associated with the classes to be recognized, e.g.\ phones in speech or notes in music, the procedure of our approach is interpretable. 
	\end{itemize}
	
	The remainder of this paper is structured as follows: In Section \ref{sec:Methods}, we introduce the basic ESN and our proposed unsupervised input weight initialization. In Section \ref{sec:Experiment1DoorStateRecognition}, we introduce, optimize and evaluate ESNs for door state classification in videos. In Section \ref{sec:Experiment2MultiDatasetEvaluation}, we present results on a wide variety of multivariate datasets. Finally, we summarize our conclusions and give an outlook to future work in Section \ref{sec:ConclusionsAndOutlook}.
	 
	\section{Methods}
	\label{sec:Methods}
	
	Here, we introduce the basic ESN and the $K$-Means algorithm, and explain how the input weights of an ESN can be initialized using the $K$-Means algorithm.
	
	\subsection{Basic Echo State Network}
	\label{subsec:BasicEchoStateNetwork}
	
	\begin{figure}[!htb]
		\centering
		\includegraphics[width=.95\columnwidth]{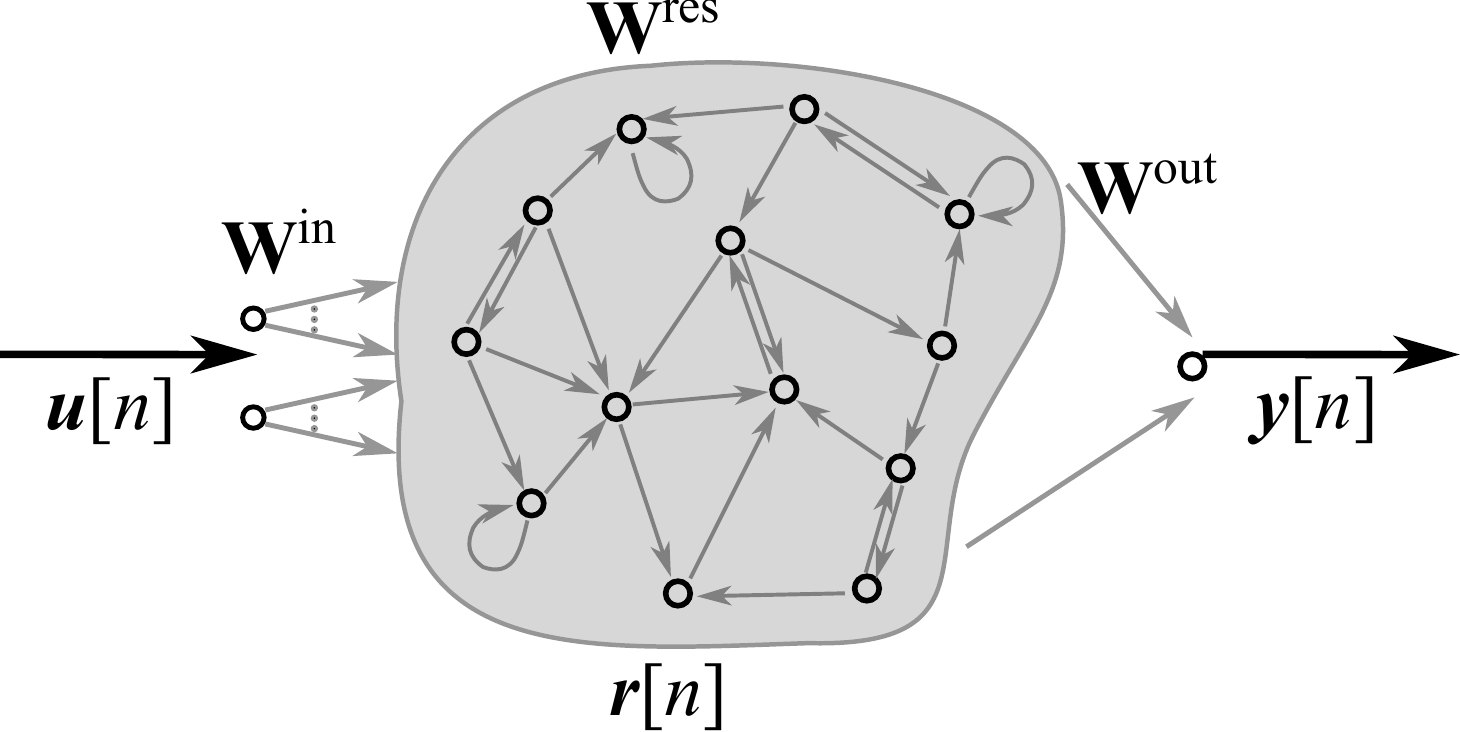}
		\caption{Main components of a basic ESN: The input features $\mathbf{u}[n]$ are fed into the reservoir using the fixed input weight matrix $\mathbf{W}^{\mathrm{in}}$. The reservoir consists of unordered neurons, sparsely inter-connected via the fixed reservoir matrix $\mathbf{W}^{\mathrm{res}}$. The output $\mathbf{y}[n]$ is a linear combination of the reservoir states $\mathbf{r}[n]$ based on the output weight matrix $\mathbf{W}^{\mathrm{out}}$, which is trained using linear regression.}
		\label{fig:algOutline}
	\end{figure}
	
	The main outline of a basic ESN is depicted in Fig.\ \ref{fig:algOutline}. The model consists of the input weights $\mathbf{W}^{\mathrm{in}}$, the reservoir weights $\mathbf{W}^{\mathrm{res}}$ and the output weights $\mathbf{W}^{\mathrm{out}}$. The input weight matrix $\mathbf{W}^{\mathrm{in}}$ has the dimension of $N^{\mathrm{res}}\times N^{\mathrm{in}}$, where $N^{\mathrm{res}}$ and $N^{\mathrm{in}}$ are the size of the reservoir and dimension of the input feature vector $\mathbf{u}[n]$ with the time index $n$, respectively. Typically, the values inside the input weight matrix are initialized randomly from a uniform distribution between $\pm 1$ and are scaled afterwards using the input scaling factor $\alpha_{\mathrm{u}}$, which is a hyper-parameter to be tuned. In~\cite{src:Jalalvand-15b}, it was shown that it is sufficient to have only a limited number of connections from the input nodes to the nodes inside the reservoir. We therefore connect each node of the reservoir to only $K^{\mathrm{in}} = \num{10}$ ($\ll N^{\mathrm{in}}$) randomly selected input features. This makes $\mathbf{W}^{\mathrm{in}}$ very sparse and feeding the feature vectors into the reservoir potentially more efficient.
	
	The reservoir weight matrix $\mathbf{W}^{\mathrm{res}}$ is a square matrix of the size $N^{\mathrm{res}}\times N^{\mathrm{res}}$. Typically, the values inside this matrix are initialized from a standard normal distribution. Similar to the input weight matrix, we connect each node inside the reservoir to a limited number of $K^{\mathrm{rec}} = \num{10}$ ($\ll N^{\mathrm{res}}$) randomly selected other nodes in the reservoir, and set the remaining weights to zero. In order to fulfil the Echo State Property (ESP) that requires that the states of all reservoir neurons need to decay in a finite time for a finite input pattern, the reservoir weight matrix is normalized by its largest absolute eigenvalue and rescaled by the spectral radius $\rho$, because it was shown in \cite{src:Jaeger-01a} that the ESP holds as long as $\rho < 1$. 
	
	Together, the input scaling factor $\alpha_{\mathrm{u}}$ and the spectral radius $\rho$ determine, how strongly the network relies on the memorized past inputs compared to the present input. These hyper-parameters need to be optimized during the training process.
	
	Every neuron inside the reservoir receives an additional constant bias input. The bias weight vector $\mathbf{w}^{\mathrm{bi}}$ with $N^{\mathrm{res}}$ entries is initialized by fixed random values from a uniform distribution between $\pm 1$ and multiplied by the hyper-parameter $\alpha_{\mathrm{b}}$. With the three weight matrices $\mathbf{W}^{\mathrm{in}}$, $\mathbf{W}^{\mathrm{res}}$ and $\mathbf{w}^{\mathrm{bi}}$ the reservoir state $\mathbf{r}[n]$ can be computed as follows:
	
	\begin{align}
	\label{eq:reservoir_state}
	\begin{split}
	\mathbf{r}[n] = & (1-\lambda)\mathbf{r}[n-1] + \\ & \lambda f_{\mathrm{res}}(\mathbf{W}^{\mathrm{in}}\mathbf{u}[n] + \mathbf{W}^{\mathrm{res}}\mathbf{r}[n-1] + \mathbf{w}^{\mathrm{bi}}) 
	\end{split}
	\end{align}

	Equation~\eqref{eq:reservoir_state} is a leaky integration of the reservoir neurons, which is equivalent to a first-order lowpass filter. Depending  on the leakage $\lambda \in (0, 1]$, a specific amount of the past reservoir state is leaked over time. Together with the spectral radius $\rho$, the leakage $\lambda$ determines the temporal memory of the reservoir. 
	
	The reservoir activation function $f_{\mathrm{res}}(\cdot)$ controls the non-linearity of the system. Conventionally, the sigmoid or $\tanh$ functions are used, because their lower and upper boundaries facilitate the reservoir states stability.  
	
	The output weight matrix $\mathbf{W}^{\mathrm{out}}$ has the dimensions $N^{\mathrm{out}}\times(N^{\mathrm{res}} + 1)$ and connects the reservoir state $\mathbf{r}[n]$, which is expanded by a constant intercept term of \num{1} for regression, to the output vector $\mathbf{y}[n]$ using Equation~\eqref{eq:readout}. 
	
	\begin{align}
	\label{eq:readout}
	\mathbf{y}[n] = \mathbf{W}^{\mathrm{out}}\mathbf{r}[n]
	\end{align}
	
	Typically, the output weight matrix is computed using ridge regression. Therefore, all reservoir states calculated for the training data are concatenated into the reservoir state collection matrix $\mathbf{R}$. As linear regression usually contains one intercept term, every reservoir state $\mathbf{r}[n]$ is expanded by a constant of 1. All desired outputs $\mathbf{d}[n]$ are collected into the output collection matrix $\mathbf{D}$. Then, $\mathbf{W}^{\mathrm{out}}$ can be computed using Equation~\eqref{eq:linearRegression}, where $\epsilon$ is the regularization parameter that needs to be tuned on a validation set.
	
	\begin{align}
	\label{eq:linearRegression}
	\mathbf{W}^{\mathrm{out}} =\left(\mathbf{R}\mathbf{R}^{\mathrm{T}} + \epsilon\mathbf{I}\right)^{-1}(\mathbf{D}\mathbf{R}^{\mathrm{T}})
	\end{align}
	
	The size of the output weight matrix determines the total number of free parameters to be trained in ESNs. Because linear regression can be obtained in closed form, ESNs are quite efficient and fast to train compared to typical deep-learning approaches.
	
	\subsection{K-Means Clustering}
	\label{subsec:K-Means-Clustering}
	
	In this work, we studied the frequently used $K$-Means algorithm \cite{src:Lloyd-82} to improve the input weight initialization of ESNs. The $K$-Means algorithm groups $N$ feature vectors (observations) $\mathbf{u}[n]$ with $N^{\mathrm{in}}$ features into $K$ clusters. Each observation is assigned to the cluster with the closest centroid, the prototype of the cluster. The basic $K$-Means algorithm aims to partition all $N$ observations into $K$ sets $S_1, S_2,\dots, S_{K}$ and thereby minimizes the within-cluster sum of squares ($\mathrm{SSE}$).
	
	\begin{align}
	\label{eq:K-MeansObjective}
	\mathrm{SSE} = \sum\limits_{k=1}^{K}\sum\limits_{\mathbf{u}[n]\in S_{k}} \|\mathbf{u}[n] - \mathbf{\mu}_{k}\|^{2}\text{ . }
	\end{align}
	
	Here, $\mathbf{\mu}_{k}$ is the centroid of the $k$-th set $S_k$, which is usually the mean of all points belonging to $S_{k}$. 
	
	In this paper, we utilized relatively large datasets. Thus, we used the fast Mini-batch $K$-Means algorithm proposed by Sculley et al.\ \cite{src:Sculley-10} to determine the $\mathbf{\mu}_{k}$ and initialized it based on \enquote{$K$-Means++} \cite{src:Arthur-07}.
	
	\subsection{Novel input weight initialization}
	\label{subsec:NovelInputWeightInitialization}
	
	In this paper, we propose to initialize the input weight matrix $\mathbf{W}^{\mathrm{in}}$ using the cluster centers $\mathbf{\mu}_{\mathrm{k}}$ determined by the $K$-Means algorithm. To understand how a feature vector is passed to the reservoir in general, we reconsider Equation \eqref{eq:reservoir_state}, which describes the computation of a new reservoir state based on the current feature vector and the previous reservoir state. For the sake of simplicity, we briefly assume a reservoir without leakage ($\lambda=1$), without any recurrent connections ($K^{rec}=0$) and with a linear activation function $f_{\mathrm{res}}$. Thus, we can simplify Equation \eqref{eq:reservoir_state} to Equations \eqref{eq:reservoir_state_memoryless} and \eqref{eq:one_reservoir_state_memoryless} for the $k^{\text{th}}$ reservoir neuron.
	
	\begin{align}
	\label{eq:reservoir_state_memoryless}
	\mathbf{r}[n] = \mathbf{W}^{\mathrm{in}}\mathbf{u}[n] 
	\end{align}
	
	\begin{align}
	\label{eq:one_reservoir_state_memoryless}
	r_{k}[n] = \sum\limits_{m=1}^{N^{\mathrm{in}}}w_{k, m}^{\mathrm{in}} u_{m}[n] = \mathbf{w}_k^{\mathrm{in}} \cdot \mathbf{u}[n]  \text{ , }
	\end{align}
	
	where $m$ is the feature index inside $\mathbf{u}[n]$, and $\mathbf{w}_k^{\mathrm{in}}$ is the $k^{\text{th}}$ row of $\mathbf{W}^{\mathrm{in}}$ with the pre-synaptic input weights for the $k^{\text{th}}$ neuron in the reservoir.
	
	This dot product is in fact closely related to the cosine similarity $S$ in Equation \eqref{eq:CosineSimilarity}. The only difference between Equations \eqref{eq:one_reservoir_state_memoryless} and \eqref{eq:CosineSimilarity} is the normalization. 
	
	\begin{align}
	\label{eq:CosineSimilarity}
	S = \dfrac {1}{||\mathbf{w}_k^{\mathrm{in}}||\;||\mathbf{u}[n]||} \; \mathbf{w}_k^{\mathrm{in}} \cdot \mathbf{u}[n]
	\end{align}
	
	The input weights of an ESN are responsible for passing feature vectors to the reservoir that consists of non-linear neurons. Due to the -- typically -- random initialization of the input weights, several linear combinations of the input features for different neurons in the reservoir can be computed. In this paper, we do not neglect this assumption, but we hypothesize that the main task of the input weights is to structure features according to their similarity. We also stick to the conventional linear regression-based training of ESNs, because this is a key-advantage of such networks. However, we would like to incorporate prior knowledge about the feature vectors, in an unsupervised fashion, so that it is \enquote{easier} for the ESN to solve a specific tasks.
	
	Thus, we propose to replace the randomly initialized input weights by the cluster centroids obtained from the $K$-means algorithm, i.e., $\mathbf{w}_k^{\mathrm{in}}=\mathbf{\mu}_{k}$. The $K$-means algorithm detects prior structure in the feature vectors, such as phones or phone transitions in speech datasets, common segments of images \cite{src:Coates-11}. This way, passing feature vectors to the ESN basically consists of computing the cosine similarity between the centroids and the feature vectors.
	
	Typically, the reservoir size $N^{\mathrm{res}}$ is increased after tuning the hyper-parameters using small ESNs. We have been shown that this final step significantly improves the classification results \cite{src:Triefenbach-10,src:Jalalvand-15b,src:Jalalvand-18,src:Steiner-20a, src:Steiner-20b,src:Steiner-20c}. However, if we would simply increase the reservoir size in our novel ESN model, we needed also to increase $K$, as each reservoir neuron represents one cluster so far. If we increase $K$ too much, we might end up with less meaningful clusters. Thus, in this paper, we propose that $K$ does not need to be equal to $N_{\mathrm{res}}$. 
	
	This has the advantage that we can increase $K$ and $N_{\mathrm{res}}$ together, until the improvement of further increasing $K$ gets low. Then, we can keep $K$ constant and add additional \enquote{zero-connections} to $\mathbf{W}^{\mathrm{in}}$. In that way, we ensure that the centroids are still representing useful information, and we reduce the computational complexity by using very sparse $\mathbf{W}^{\mathrm{in}}$ in case of large reservoirs. By padding the new input weights with a lot of \enquote{zero-connections}, we specifically limit the amount of neurons in the reservoir that receive input features. This can be compared with a cortical column in the brain that also mainly consists of recurrent connections and in which only a part of the neurons directly receives input information~\cite{src:Maass-11}. Thus, very large $K$-Means-based ESNs are even getting biologically plausible.
	
\begin{figure}[!htb]
		\centering
		\subfloat[$K$-Means-based ESN when $K=N^{\mathrm{res}}=3$.]{\includegraphics[width=.95\columnwidth]{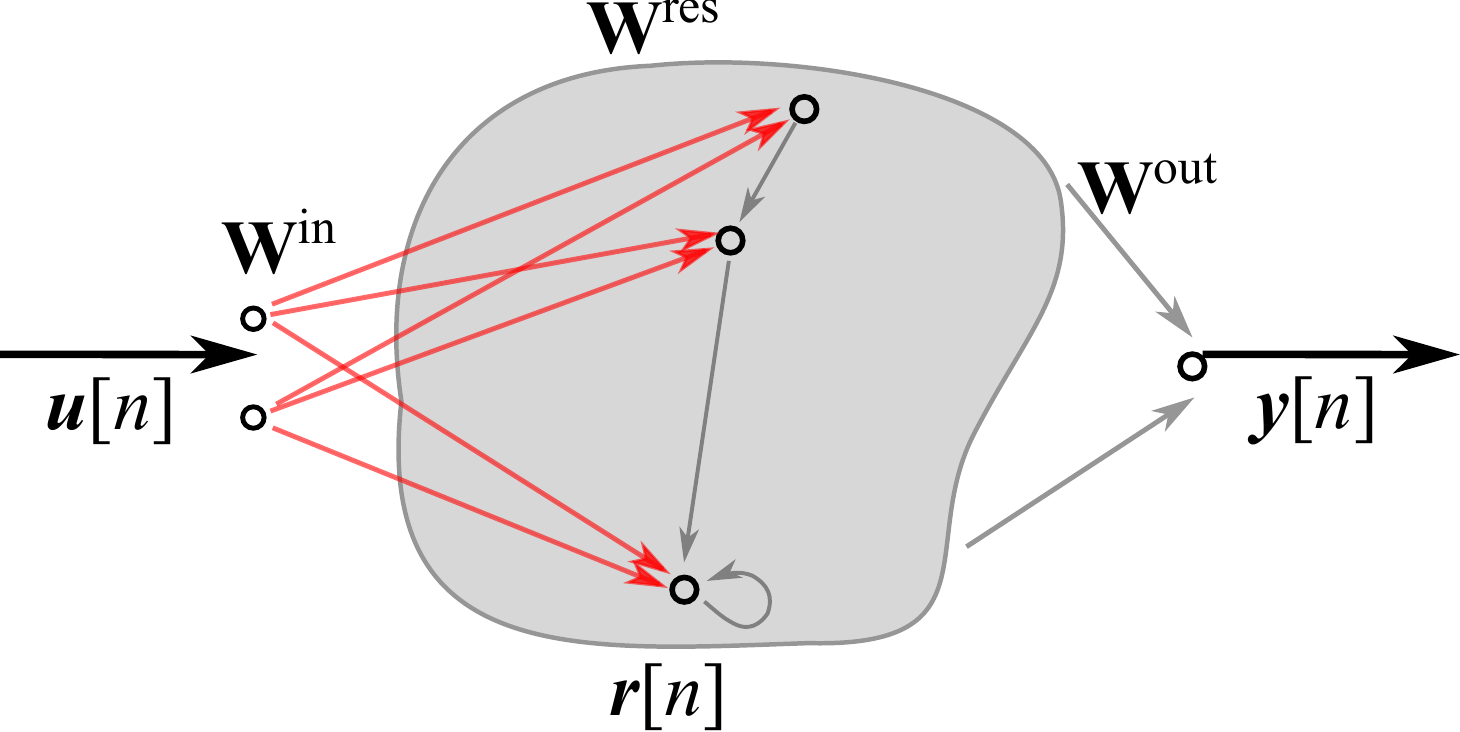}\label{fig:algOutlineKMeansSmall}}

		\subfloat[$K$-Means-based ESN when $K<N^{\mathrm{res}}$.]{\includegraphics[width=.95\columnwidth]{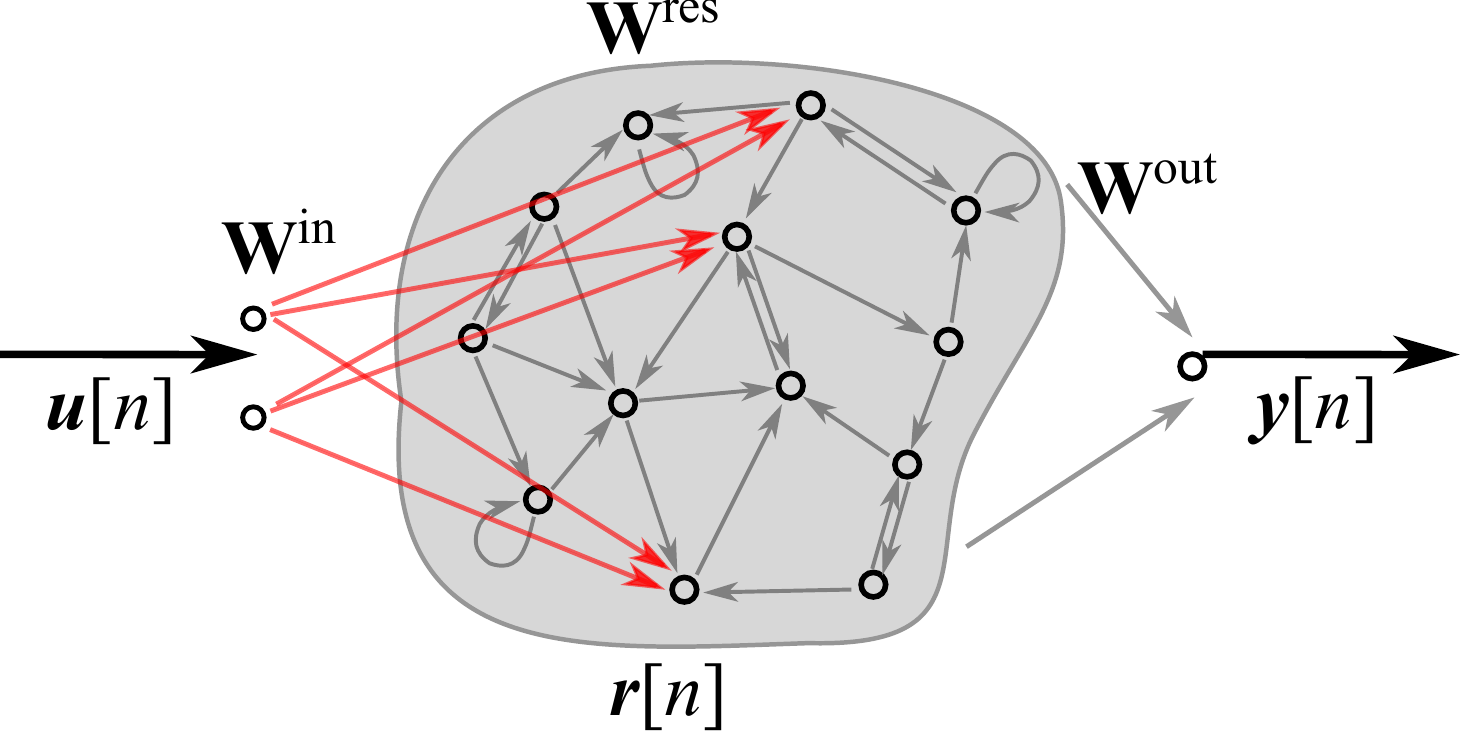}\label{fig:algOutlineKMeansLarge}}

		\caption{Example ESN architectures for $K=N^{\mathrm{res}}=3$ (Figure \ref{fig:algOutlineKMeansSmall}) and for $K=3$ and $N^{\mathrm{res}}=13$ (Figure \ref{fig:algOutlineKMeansLarge}). Only the non-zero parts of $\mathbf{W}^{\mathrm{in}}$ are visualized in red color. For the sparse KM-ESN, only a few neurons receive input data.}
		\label{fig:algOutlineKMeansSmallLarge}
	\end{figure}

	Figure~\ref{fig:algOutlineKMeansSmallLarge} visualizes two very simple examples for the proposed $K$-Means-based ESN architectures. Fig.\ \ref{fig:algOutlineKMeansSmall} shows the case $K=N^{\mathrm{res}}=3$, where the number of centroids is equal to the reservoir size. Figure~\ref{fig:algOutlineKMeansLarge} shows the case $K=3$ and $N^{\mathrm{res}}=13$, where the number of centroids is much smaller than the reservoir size.  As indicated by the red arrows, only a very small amount of the neurons inside the reservoir receive information directly from the input features, whilst most of the neurons are only connected to other neurons inside the reservoir.
	
	In the remainder of this paper, we refer to the basic ESN with randomly and sparsely initialized input weights as \enquote{basic ESN} and to the ESN with $K$-Means-based initialized input weights as \enquote{KM-ESN}. If $N^{\mathrm{res}}>K$, we call it \enquote{sparse KM-ESN}.
	
	\section{Experiment 1: Door State recognition}
    \label{sec:Experiment1DoorStateRecognition}
    
    In the first experiment, we consider a frame-level classification task, namely, event detection in door surveillance systems. The task is to continuously classify the status of a door that can be open, closed or half-open from a low resolution camera sensor. Using a large-scale dataset, we illustrate the impact of the $K$-Means clustering on the hyper-parameters.
	
    \subsection{Dataset}
    \label{subsec:Experiment1Dataset}
    
    We used the publicly available dataset \cite{src:Jalalvand-15a} that contains recordings of a low-resolution camera (Avago Technologies ADNS-3080 mouse sensor) set up in front of a door. The resolution was $\num{30}\times\num{30}$ pixels with a frame rate of \num{90} frames per second. The dataset has more than \num{830000} frames in total. For each frame, the label (0 for the closed, 1 for the half-opened and 2 for the opened door, respectively) was semi-automatically generated using magnetic sensors that were placed in the middle of the door and close to the door hinge. In the dataset, three movies of different lengths are included, where the camera position in each movie was slightly displaced to introduce more variable input features.
    
    The dataset does not have any default split into training, validation and test sets. We prepared the dataset as follows: Each movie was split into consecutive sequences with a length of $\approx 1$ minute. The first half of each movie was used as training set ($N$) and the latter one as the test set. To optimize the hyper-parameters, we only used the first movie, whose training set was partitioned in five folds for cross validation. We sequentially optimized the hyper-parameters according to \cite{src:Steiner-20a}.
    
    \subsection{Feature Extraction}
    \label{subsec:Experiment1FeatureExtraction}
    
    Following \cite{src:Jalalvand-15a}, we converted each frame with $\num{30}\times\num{30}$ pixels to a vector with \num{900} elements and did not consider further feature reduction techniques. Since the pixel values were integers between 0 and 255 (grayscale values), we divided each value by 255 to obtain values between 0 and 1. We did not do any further pre-processing steps and directly used the rescaled vectors as input for the ESN models. The training set contained $\approx 76$ minutes of data leading to $N_{\mathrm{samples}} = \num{415620}$ frames in total. 
    
    \subsection{Target preparation and readout post-processing}
    \label{subsec:Experiment1TargetPreparation}
    
    In this task, we have three binary outputs, one for each door state (close, half-open and open). We converted the integer label of each frame in a three-dimensional output with one-hot encoded targets, where only the output indexed by the integer label is \num{1}. 
    
    During inference, the output with the highest value in every frame indicated the current state of the door.
    
    \subsection{Measurements}
    \label{subsec:Experiment1Measurements}
    
    To optimize the hyper-parameters, we used the mean squared error ($\mathrm{MSE}$) between the one-hot encoded targets and the computed outputs. 
    
    To report the final performance, the frame-level error rate $\mathrm{FER}$ (portion of frames assigned to the wrong class, Eq.\ \ref{eq:FER}) was used. 
    
    \begin{align}
    	\label{eq:FER}
    	\mathrm{FER} = \dfrac{\mathrm{N}_{\mathrm{error}}}{\mathrm{N}_{\mathrm{frames}}}\text{ , }
    \end{align}
    
    where $\mathrm{N}_{\mathrm{error}}$ and $\mathrm{N}_{\mathrm{frames}}$ were the misclassified frames and the total number of frames, respectively. 
    
    \subsection{Number of centroids}
    \label{subsec:Experiment1OptimizeKMeans}
    
\begin{figure}[!htb]
    	\centering
    	\input{figs/results/video_dataset/kmeans_opt_video.tex}
    	\caption{Summed Squared Error ($\mathrm{SSE}$) for different numbers of centroids. A fast decrease was observed until $K\approx\num{20}$. Afterwards, $\mathrm{SSE}$ still continued to decrease more slowly.}
    	\label{fig:kmeans_minibatch_elbow_curve}
    \end{figure}
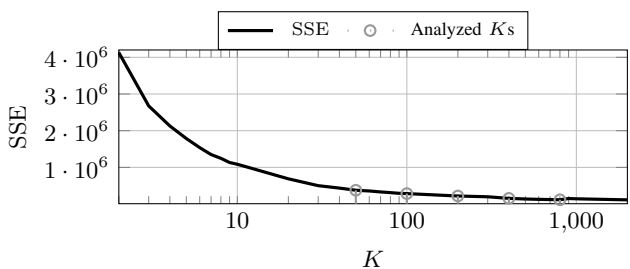

    The key parameter of the $K$-means algorithm is $K$, the number of clusters to be used. This is strongly task-dependent and there exists no general solution for this optimization problem. One way to determine the number of centroids is to observe the summed squared error $\mathrm{SSE}$ for different $K$ and to search for the point when the slope of the $\mathrm{SSE}$ gets less steep. From Figure \ref{fig:kmeans_minibatch_elbow_curve}, where the summed squared error $\mathrm{SSE}$ is visualized over $K$ for the training set, we observed that the $\mathrm{SSE}$ decreased quite fast until $K\approx \num{50}$. Afterwards, SSE still continued to decrease but with a slower pace.
    
    The low effective number of clusters that is able to cluster the dataset to some extent is in line with the nature of this dataset. Given that the camera was fixed in each movie and that it recorded a limited set of interactions between humans and the door, the majority of all pixels were more or less constant over time. Although different persons interacted with the door, the resolution of the camera is rather low so that it mostly captures the overall shape of each person. Since the dataset is rather noisy, it was furthermore difficult for the camera to record objects (e.g.\ chairs) that the people carried into or out of the room. During the reservoir hyper-parameter optimization, we first of all fixed the number of centroids to $K=\num{50}$ and later increased it together with the reservoir size. 

    \subsection{Optimization of the hyper-parameters of the ESNs}
    \label{subsec:Experiment1OptimizatiofTheESNs}
    
\begin{figure}[!htb]
    	\centering
    	\subfloat[Basic ESN]{\includegraphics{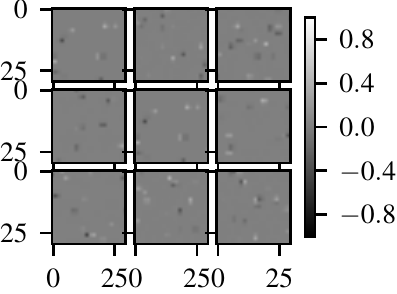}\label{subfig:Basic_ESN_some_Input_Weights}}\quad
    	\subfloat[KM-ESN]{\includegraphics{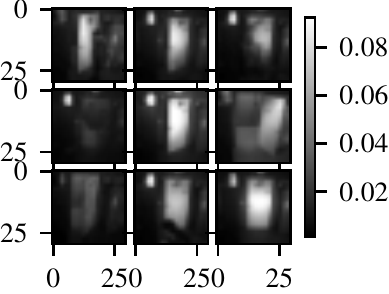}\label{subfig:KM_ESN_some_Input_Weights}}
    	\caption{Input weights for both ESN models. In case of the basic ESN (Fig.\  \ref{subfig:Basic_ESN_some_Input_Weights}), every neuron received $\approx\num{10}$ randomly selected input features and thus has $\approx\num{890}$ zero connections. The remaining values are uniformly distributed between $\pm\num{1}$. In case of the KM-ESN (Fig.\ \ref{subfig:KM_ESN_some_Input_Weights}), the weights were learned from the training dataset and represent basically mean values of various images.}
    	\label{fig:ESN_Input_Weights}
    \end{figure}

    Before optimizing the hyper-parameters of the ESN models, we compared the input weights of the basic ESN and of the KM-ESN. Therefore, the input weights to nine randomly selected reservoir neurons are visualized in Figure \ref{fig:ESN_Input_Weights} as follows: Each neuron has a \num{900}-dimensional input weight vector. We have reshaped them into a $\num{30}\times\num{30}$ images, since the input weights connect exactly one input image of the same size to the particular neuron. The following differences can be observed:
    
    \begin{itemize}
        \item The input weights of the basic ESN (Fig.\ \ref{subfig:Basic_ESN_some_Input_Weights}) are very sparse (\SI{0.06}{\percent} non-zero values) and have uniformly distributed non-zero values between $\pm1$. 
        \item Figure \ref{subfig:Basic_ESN_some_Input_Weights} shows that images are randomly fed in the basic ESN.
        \item The KM-ESN (Fig.\ \ref{subfig:KM_ESN_some_Input_Weights}) is dense and the values are approximately distributed between \num{0} and \num{0.25}. This sparseness is caused by the dark regions in the frames.  
        \item Fig.\ \ref{subfig:KM_ESN_some_Input_Weights} shows that the images are fed in the KM-ESN by the correlation with prototype images learned by the $K$-Means algorithm.
    \end{itemize}
    
    These observations can be explained by the different ways of initializing the basic ESN and the KM-ESN. For the first one, we partially followed \cite{src:Jalalvand-15a}, where each neuron received only $K^{\mathrm{in}}=\num{5}$ randomly chosen inputs. In contrast to \cite{src:Jalalvand-15a}, we increased $K^{\mathrm{in}}$ to \num{10} as we already know that the level of sparseness does not have a significant impact on the performance \cite{src:Jalalvand-15a}. The other observations show that the $K$-Means algorithm learned exactly what we expected it to learn -- average images for each cluster. Since the weights of the basic ESN and the KM-ESN are different, we expect the two models to behave differently in the remaining experiment. 
    
    The last observation is the most interesting one, since it perfectly visualizes that the $K$-Means algorithm has learned different average images from the training dataset. Thus, different opening phases of the door or silhouettes of people in the room or even the displacement of the camera due to different positions can be observed in Fig.\ \ref{subfig:KM_ESN_some_Input_Weights}.  
    
    Since the input weights of the KM-ESN are initialized using a purely data-driven approach, the value range strongly depends on the value range of the features. In case of the video dataset, the values are bounded between \num{0} and \num{1}, hence, the values of the input weights of the KM-ESN are also non-negative. For the following hyper-parameter optimization, the different distributions of the input weights will lead to significant differences between the hyper-parameters of the basic ESN and of the KM-ESN, especially for the input scaling and spectral radius.

    In order to optimize the hyper-parameters of both basic and KM-ESN, we followed the sequential optimization approach introduced in \cite{src:Steiner-20a,src:Steiner-20b}. A similar outline was recently published in \cite{src:Hinaut-21}. In our preliminary experiments, we compared this sequential method with a fully randomized optimization by jointly exploring the entire hyper-parameter space. We found that the sequential optimization required fewer search steps and led to a lower loss function. Therefore, in the following, we use the sequential optimization process.
    
    We began with a memory-less ESN ($\rho = 0$) with \num{50} reservoir neurons. Particularly, we fixed the leakage $\lambda=1$ and removed the constant bias term by setting $\alpha_{bi} = 0$.
    
    Then, we jointly optimized $\alpha_{\mathrm{u}}$ and $\rho$ using a random search with \num{200} iterations. The values for $\alpha_{\mathrm{u}}$ were drawn from a uniform distribution between \num{0.01} and \num{1.0}, and the values for $\rho$ from a uniform distribution between \num{0} and \num{2}. 
    
\begin{figure}[!htb]
    	\centering
    	\subfloat[Basic ESN]{\includegraphics[height=.34\columnwidth, trim=0 0 40.5 0, clip]{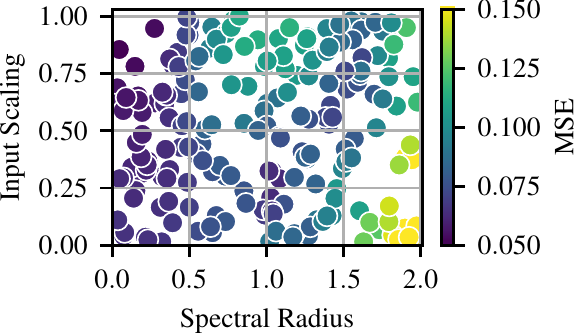}\label{subfig:Basic_ESN_IS_SR}}\quad
    	\subfloat[KM-ESN]{\includegraphics[height=.34\columnwidth, trim=27 0 0 0, clip]{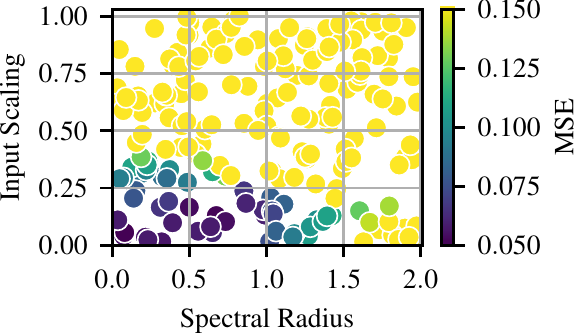}\label{subfig:KM_ESN_IS_SR}}
    	\caption{$\mathrm{MSE}$ after the joint optimization of the input scaling $\alpha_{\mathrm{u}}$ and the spectral radius $\rho$. The overall behaviour of the basic ESN and the KM-ESN is similar. The input scaling of the basic ESN (\num{0.85}) is very large compared to the one of the KM-ESN (\num{0.05}).}
    	\label{fig:ESN_IS_SR}
    \end{figure}

    The entire search space after 5-fold cross validation is depicted in Fig.\ \ref{fig:ESN_IS_SR}. To better present the results, we have visualized them using the expression $\min(\num{0.15}, \mathrm{MSE})$. We achieved the lowest validation MSE with $(\alpha_{\mathrm{u}}, \rho)=(\num{0.85}, \num{0.05})$ for the basic ESN and $(\alpha_{\mathrm{u}}, \rho)=(\num{0.05}, \num{0.08})$ for the KM-ESN. Comparing the optimization results in Fig.\ \ref{fig:ESN_IS_SR}, the general behavior of these basic and KM-ESNs is different. In particular, there are two marginal differences: The input scaling is lower in case of the KM-ESN, whereas the spectral radii are similar. The area of the best hyper-parameters in case of the KM-ESN is much smaller than the area in case of a basic ESN. 
    
    The differences in the input scaling values are caused by the different of input weights of the basic ESN and of the KM-ESN. While most of the values in the input weights of the basic ESN are zero, the remaining values need to strongly activate the reservoir neurons, which is achieved with a large input scaling. The KM-ESN in contrast, has learned prototype images. If the input image is very similar to a prototype, the cross correlation between the input image and the prototype is high, leading to a strong activation of the associated neuron. Thus, despite the smaller absolute values of the input weights in the KM-ESN, a significantly smaller input scaling value is required. Since the ratio between input scaling and spectral radius is almost 1 in case of the KM-ESN, it relies more on a combination of current and past information than the basic ESN, which mostly benefited from the current input.
    
\begin{figure}[!htb]
    	\centering
    	\subfloat[Basic ESN]{\includegraphics[height=.34\columnwidth]{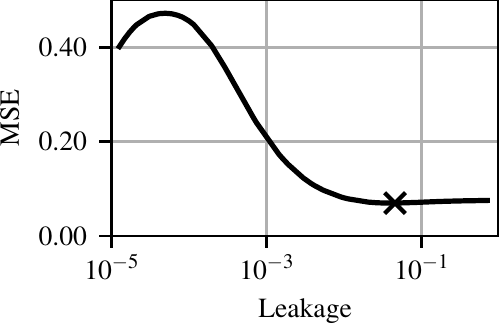}\label{subfig:Basic_ESN_LR}}\quad
    	\subfloat[KM-ESN]{\includegraphics[height=.34\columnwidth, trim=26 0 0 0, clip]{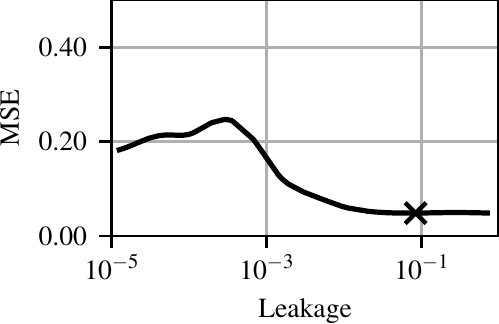}\label{subfig:KM_ESN_LR}}
    	\caption{$\mathrm{MSE}$ after the optimization of the leakage $\lambda$. The global behaviour of the basic ESN and the KM-ESN is similar. As indicated by the crosses, the optimized leakage values of the basic ESN and of the KM-ESN are \num{0.05} and \num{0.08}, respectively.}
    	\label{fig:leakage}
    \end{figure}

    The next hyper-parameter to be optimized was the leakage $\lambda$. Again, we used a random search to optimize this hyper-parameter for the basic ESN and for the KM-ESN. This time, we used a logarithmic uniform distribution between \num{1e-5} and \num{1}, because we expected that a large leakage, i.e., very small $\lambda$ is required for the ESN to make the decision based on a wide range of memory. The results in Fig.\ \ref{fig:leakage} show that the leakage has a strong impact on the final performance. Again, the global behaviour of the basic ESN and the KM-ESN was similar. The final values were \num{0.05} and \num{0.08} for the basic ESN and for the KM-ESN, respectively.
    
    The low values for $\lambda$ are reasonable for this task, since the dataset is noisy and the outputs need to be constant for a longer time, especially for long phases with an opened or closed door. Thus, a small $\lambda$ is desired that acts as a first-order lowpass filter \cite{src:Jaeger-07} and smooths the reservoir states.
    
\begin{figure}[!htb]
    	\centering
    	\subfloat[Basic ESN]{\includegraphics[height=.33\columnwidth]{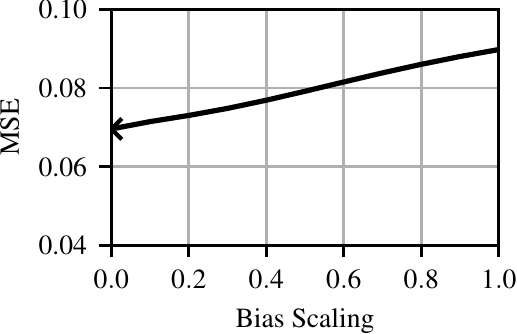}\label{subfig:Basic_ESN_BS}}\quad
    	\subfloat[KM-ESN]{\includegraphics[height=.33\columnwidth, trim=26 0 0 0, clip]{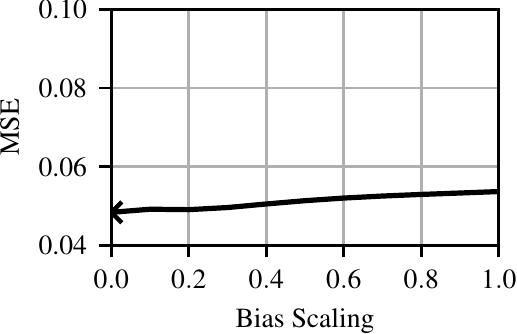}\label{subfig:KM_ESN_BS}}
    	\caption{$\mathrm{MSE}$ after the optimization of the bias scaling $\alpha_{\mathrm{bi}}$. The impact of bias scaling is rather small compared to the impact of the previous hyper-parameters. As indicated by the crosses, the best performance was achieved with $\alpha_{\mathrm{bi}} = 0$ in both cases.}
    	\label{fig:bias_scaling}
    \end{figure}

    The last hyper-parameter to be optimized was the bias scaling factor $\alpha_{\mathrm{bi}}$ that controls the influence of a constant bias input to each reservoir neuron. In general, the bias scaling has a minor impact on the final ESN performance. Thus, we simplified the optimization scheme here and evaluated values from \num{0} to \num{1} with a step of \num{0.1} to optimize this parameter. Fig.\ \ref{fig:bias_scaling} shows that the impact of the bias term is indeed small and that large bias inputs even decreased the performance of the ESN models. The basic ESN as well as the KM-ESN did not need any bias at all.

    \subsection{Impact of the reservoir size}
    \label{subsec:Experiment1ImpactOfTheReservoirSize}
    
    As mentioned in the introduction, ESNs typically benefit from increasing the reservoir size after fixing the other hyper-parameters. At the same time, we were reluctant to add to the complexity of the $K$-Means model by increasing $K$ as much as $N_{res}$. Therefore, we only increased $K$ and $N_{res}$ up to \num{200} and from that point we only increased the reservoir size while $K=\num{200}$. This means that the input layers of the ESNs were fully connected until $N_{\mathrm{res}}=\num{200}$ and for all the larger models, the input features were connected to only \num{200} reservoir nodes (i.e., sparse input connections). We also repeated the hyper-parameter optimization for these sparsely connected KM-ESNs. 
    
    The optimized hyper-parameters for the new initialized ESNs are summarized in Table \ref{tab:newHyperParams}. Since introducing neurons in the sparse reservoir that did not receive any input information strongly changes the overall system behaviour, the optimal hyper-parameters have changed. In particular, it is interesting that the spectral radius strongly increased by a factor of ten while input scaling increased only by a factor of three. This means that especially the sparse KM-ESN not only uses the current input to compute the output but also needs the memory provided by the recurrent connections. Less smoothing by the leaky integration is required now.

    \begin{table}[!htb]
    	\renewcommand{\arraystretch}{1.3}
    	\caption{Optimized hyper-parameters for the basic ESN, small KM-ESN ($N_{res}<\num{200}$ and dense input layer) and large KM-ESN ($N_{res}>\num{200}$ and sparse input layer) after 5-fold cross validation.}
    	\label{tab:newHyperParams}
    	\centering
    	\begin{tabular}{llll}
    		\toprule
    		& Basic ESN & \multicolumn{2}{c}{KM-ESN} \\
    		Parameter &  & dense & sparse \\
    		\midrule
    		Input scaling $\alpha_{\mathrm{u}}$ & \num{0.85} & \num{0.06} & \num{0.13} \\
    		Spectral radius $\rho$ & \num{0.05} & \num{0.08} & \num{0.99} \\
    		Leakage $\lambda$ & \num{0.04} & \num{0.08} & \num{0.35} \\
    		Bias scaling $\alpha_{\mathrm{bi}}$ & \num{0} & \num{0} & \num{0} \\
    		\midrule
    		Regularization $\epsilon$ & \num{13e-4} & \num{6e-4} & \num{3e-4} \\
    		\bottomrule
    	\end{tabular}
    \end{table}
    
\begin{figure}[!htb]
    	\centering
    	\input{figs/results/video_dataset/video_final_score.tex}
    	\caption{Mean, minimum and maximum Frame Error Rate ($\mathrm{FER}$) for different reservoir sizes after ten random initializations. The KM-ESN always performed equally well or better than the basic ESN for all reservoir sizes. The vertical bars indicate the minimum and maximum $\mathrm{FER}$ of the respective models.}
    	\label{fig:video_final_result}
    \end{figure}
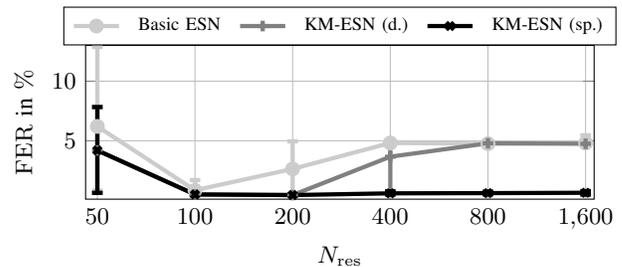

    In Figure \ref{fig:video_final_result}, the final $\mathrm{FER}$ computed on the test set for different ESN architectures are visualized. Overall, for both basic ESN and KM-ESN, only a small reservoir with \num{100} neurons was required to achieve a reasonable performance that was clearly below \SI{1}{\percent} $\mathrm{FER}$. In particular, the best performing basic ESN with \num{100} neurons achieved an average $\mathrm{FER}$ of \SI{0.85}{\percent} and the same KM-ESN an average $\mathrm{FER}$ of \SI{0.50}{\percent}. In case of reservoirs with more than \num{100} neurons, we noticed that the performance of the basic ESN strongly decreased with an $\mathrm{FER}$ of about $\SI{5}{\percent}$. The same effect occurred in case of the dense KM-ESN with more than \num{200} neurons. Reconsidering the $\mathrm{SSE}$ in Fig.\ \ref{fig:kmeans_minibatch_elbow_curve}, we noticed that it almost stopped decreasing with more than $\num{200}$ neurons. In case of large basic ESNs, it is likely that many static pixels are selected by the input weights, and in case of large dense KM-ESNs, it is likely that a lot of centroids are close to each other. This, in turn, means that large KM-ESNs receive a lot of input information in an almost equivalent way, which is counterproductive since ESNs in general benefit from diverse input. By switching to the sparse KM-ESN when increasing the reservoir size beyond \num{200} neurons, we restricted the number of centroids and thus did not split up meaningful clusters in too many small subclusters. In addition, we boosted the impact of the recurrent connections, since the number of parameters in $W^{\mathrm{res}}$ increases quadratically with $N^{\mathrm{res}}$. From Fig.\ \ref{fig:video_final_result}, we can conclude that the performance does not decrease in case of large sparse KM-ESNs and further improved to $\mathrm{FER}=\SI{0.44}{\percent}$ for $K=\num{200}$ and $N^{\mathrm{res}}=400$. Reconsidering Table \ref{tab:newHyperParams}, we can say that large reservoirs benefit from the memory incorporated by means of the high spectral radius. For $N^{\mathrm{res}}=\num{1600}$, the $\mathrm{FER}$ slightly increased towards \SI{0.64}{\percent}. Overall, the KM-ESN (regardless dense or sparse) was always more robust against ten different random initalizations with only one exception: the dense KM-ESN with \num{400} neurons, when the performance got more similar to the basic ESN.

    \subsection{Computational complexity}
    \label{subsec:Experiment1ComputationalComplexity}
    
    According to \cite{src:Jalalvand-15b}, passing data through the ESN can be decomposed in a series of actions that include computing the reservoir states, updating the correlation matrices, matrix inversion and output weight computation. In case of the KM-ESN, the $K$-Means training is an additional initial step.
    
    The complexity of all steps are summarized in Table \ref{tab:computationalComplexity} and show that the $K$-Means algorithm adds to the complexity, but needs to be performed only once before the  hyper-parameter optimization.
    
    \begin{table}[!htb]
    	\renewcommand{\arraystretch}{1.3}
    	\caption{Computational complexity to train the basic ESN and the KM-ESN. The $K$-Means algorithm increases the complexity but needs to be performed only once before the hyper-parameter optimization.}
    	\label{tab:computationalComplexity}
    	\centering
        \begin{tabular}{ll}
        \toprule
        Action & Complexity \\
        \midrule
        $K$-Means \cite{src:Pakhira-14,src:Bachem-16} & $\mathcal{O}(N^{\mathrm{in}} + N_{\mathrm{samples}}KN^{\mathrm{in}})$ \\
        \midrule
        Compute $\mathbf{R}$ & $\mathcal{O}(N_{\mathrm{samples}}N^{\mathrm{res}})$ \\
        Update $\mathbf{R}\mathbf{R}^{\mathrm{T}}$ & $\mathcal{O}(N_{\mathrm{samples}}(N^{\mathrm{res}})^{2})$ \\
        Update $\mathbf{D}\mathbf{R}^{\mathrm{T}}$ & $\mathcal{O}(N_{\mathrm{samples}}N^{\mathrm{res}}N^{\mathrm{out}})$ \\
        Invert $\left(\mathbf{R}\mathbf{R}^{\mathrm{T}} + \epsilon\mathbf{I}\right)$ & $\mathcal{O}((N^{\mathrm{res}})^{3})$ \\
        Compute $\mathbf{W}^{\mathrm{out}}$ & $\mathcal{O}((N^{\mathrm{res}})^{2}N^{\mathrm{out}})$ \\
        \bottomrule
        \end{tabular}
    \end{table}
	
	\section{Experiment 2: Multi-dataset evaluation}
	\label{sec:Experiment2MultiDatasetEvaluation}
	
	In this section, we focus on evaluating the KM-ESN on a large variety of datasets with different characteristics, such as dataset size, feature vector size, sequence length and data type. 

	\begin{table*}[!htb]
	    \caption{Details about the datasets for the multi-dataset evaluation with the number of input features ($N^{\mathrm{in}}$), number of outputs ($\mathrm{N^{\mathrm{out}}}$), mean, minimum and maximum sequence duration ($T_{\mathrm{mean}}$, $T_{\min}$ and $T_{\max}$), the entire number of feature vectors (observations) in the training set and test set ($N_{\mathrm{samples},\mathrm{train}}$, $N_{\mathrm{samples},\mathrm{test}}$), and the number of sequences in the training and test set ( \#Train, \#Test).}
	    \label{tab:MultiDatasets}
	    \centering
	    \begin{tabular}{lllllllllll}
	    \toprule
	    Dataset & Abbreviation & $N^{\mathrm{in}}$ & \#Train & \#Test & $N^{\mathrm{out}}$ & $T_\mathrm{mean}$ & $T_{\min}$ & $T_{\max}$ & $N_{\mathrm{samples},\mathrm{train}}$ & $N_{\mathrm{samples},\mathrm{test}}$ \\
	    \midrule
	    Swedish Leaf & SWE & \num{1} & \num{500} & \num{625} & \num{15} & \num{128} & \num{128} & \num{128} & \num{64000} & \num{80000} \\
	    Chlorine Concentration & CHLO & \num{1} & \num{367} & \num{3840} & \num{3} & \num{166} & \num{166} & \num{166} & \num{77522} & \num{637440} \\
        DistalPhalanxTW & PHAL & \num{1} & \num{400} & \num{139} & \num{3} & \num{80} & \num{80} & \num{80} & \num{32000} & \num{11120} \\
	    \midrule
	    ECG & ECG & 2 & 100 & 100 & 2 & \num{89.5} & \num{39} & \num{152} & \num{9014} & \num{8884} \\
	    Libras & LIB & \num{2} & \num{180} & \num{180} & \num{15} & \num{45} & \num{45} & \num{45} & \num{8100} & \num{8100} \\
	    \midrule
	    Character Trajectories & CHAR & \num{3} & \num{300} & \num{2558} & \num{20} & \num{120} & \num{109} & \num{205} & \num{36070} & \num{306869} \\
	    uWave & UWAV & \num{3} & \num{200} & \num{427} & \num{8} & \num{315} & \num{315} & \num{315} & \num{63000} & \num{134820} \\
	    \midrule
	    NetFlow & NET & \num{4} & \num{803} & \num{534} & \num{13} & \num{230.7} & \num{50} & \num{994} & \num{182881} & \num{125506} \\
	    \midrule
	    Wafer & WAF & \num{6} & \num{298} & \num{896} & \num{2} & \num{136.8} & \num{104} & \num{198} & \num{40833} & \num{122450} \\
	    Robot Failures & ROBOT & \num{6} & \num{100} & \num{64} & \num{4} & \num{14.8} & \num{12} & \num{15} & \num{1476} & \num{950} \\
	    \midrule
	    Japanese Vowels & JPVOW & \num{12} & \num{270} & \num{370} & \num{9} & \num{15.6} & \num{7} & \num{29} & \num{4274} & \num{5687} \\
	    \midrule
	    Arabian Digits & ARAB & \num{13} & \num{6600} & \num{2200} & \num{10} & \num{39.8} & \num{4} & \num{93} & \num{263256} & \num{87063} \\
	    \midrule
	    Auslan & AUS & \num{22} & \num{1140} & \num{1425} & \num{95} & \num{57.3} & \num{45} & \num{136} & \num{63371} & \num{83578} \\
	    \midrule
	    CMU subject \num{16} & CMU & \num{62} & \num{16} & \num{10} & \num{2} & \num{305.0} & \num{127} & \num{580} & \num{8462} & \num{9229} \\
	    Kick vs.\ Punch & KICK & \num{62} & \num{16} & \num{10} & \num{2} & \num{426.7} & \num{274} & \num{841} & \num{6413} & \num{4682} \\
	    Walk vs.\ Run & WALK & \num{62} & \num{28} & \num{16} & \num{2} & \num{367.9} & \num{128} & \num{1918} & \num{10926} & \num{5261} \\
	    \midrule
	    PEMS & PEMS & \num{963} & \num{267} & \num{173} & \num{7} & \num{144} & \num{144} & \num{144} & \num{38448} & \num{24912} \\
	    \bottomrule
	    \end{tabular}
	\end{table*}

    \subsection{Datasets}
	\label{subsec:Experiment2Datasets}
	
	We used exactly the same datasets as in \cite{src:Bianchi-20}, which were provided in the accompanying Github repository\footnote{\href{https://github.com/FilippoMB/Time-series-classification-and-clustering-with-Reservoir-Computing}{https://github.com/FilippoMB/Time-series-classification-and-clustering-with-Reservoir-Computing}}. Statistics about the datasets are summarized in Table \ref{tab:MultiDatasets} and show the diversity of the tasks such as single and multi-input time-series as well as binary classification and multi-class tasks.
	
	The datasets are by default split in training and test subsets. As before, for hyper-parameter optimization, we partitioned the training set in five folds for cross validation and then again sequentially optimized the hyper-parameters as in \cite{src:Steiner-20a} and in the previous experiment. Since the datasets \enquote{CMU subject 16}, \enquote{Kick vs.\ Punch} and \enquote{Walk vs.\ Run} contained only very few training sequences, we used 3-fold cross validation to optimize the hyper-parameters for these tasks. 

	\subsection{Feature extraction}
	\label{subsec:Experiment2FeatureExtraction}
	
	Most of the datasets were already pre-processed to some extent. For all datasets except for NetFlow, we subtracted the mean value from each feature and normalized it to unitary variance. 
	
	The NetFlow dataset was almost binary and the proposed normalization was not applicable. Instead, we simply rescaled each feature to the range between \num{0} and \num{1}.
	
	\subsection{Target preparation and readout post-processing}
	\label{subsec:Experiment2TargetPreparation}
    
    For all datasets, we have binary outputs, one for each class in the particular dataset. For each dataset, the target outputs (classes) were one-hot encoded across the entire sequence (\num{0} for the inactive classes and \num{1} for the active class). 

    During inference, the class scores were obtained by accumulating the class readouts over time. The \emph{recognised} class is determined as the class with the highest accumulated score over time.
	
	\subsection{Measurements}
	\label{subsec:Experiment2Measurements}
	
	To measure the overall classification results and to optimize the hyper-parameters, the Classification Error Rate $\mathrm{CER}$ \eqref{eq:CER} was used.
	
	\begin{align}
	\label{eq:CER}
	\mathrm{CER} = \dfrac{\mathrm{N}_{\mathrm{error}}}{\mathrm{N}_{\mathrm{seq}}}\text{ , }
	\end{align}
	
	where $\mathrm{N}_{\mathrm{error}}$ and $\mathrm{N}_{\mathrm{seq}}$ were the number of incorrect classified sequences and the overall number of sequences, respectively. 
	
	\subsection{Number of centroids}
	\label{subsec:Experiment2OptimizeKMeans}
	
	As in the previous experiment, we optimized the number of clusters $K$ of the $K$-Means algorithm. For each dataset, we computed the $\mathrm{SSE}$ for different $K$ and evaluated the values \numlist{50;100;200;400;800;1600}. In some datasets, e.g.\ the Auslan (AUS) and the CMU subject 16 (CMU) datasets, we observed that the performance decreased or stagnated when increasing $K$ too much. In that case, we stopped increasing $K$ as soon as the performance dropped and instead switched to a sparse KM-ESN when further enlarging the reservoir size. Since the ROBOT dataset contained less than \num{1600} samples, we did not evaluate $K=\num{1600}$.

	\subsection{Optimization of the hyper-parameters of the ESNs}
	\label{subsec:Experiment2OptimizationtheESNs}
	
	The optimization of the ESN models followed the same strategy as in Sec.\ \ref{subsec:Experiment1OptimizatiofTheESNs} with one exception: Instead of starting with a default leakage of \num{1.0}, we chose \num{0.1}, because the target outputs were constant across the entire sequence. Thus, we expected a lower leakage. The subsequent optimization procedure was then the sequential optimization of the hyper-parameters, during which we minimized the cross validation $\mathrm{MSE}$. 
	
	The hyper-parameters with the lowest cross validation $\mathrm{MSE}$ were then used as the final hyper-parameters. During the optimization, we fixed the reservoir size to \num{50} neurons.
	
	As discussed for the previous experiment, the hyper-parameters significantly influence the performance of the ESN and they need to be tuned task-dependently. Thus, in contrast to \cite{src:Bianchi-20}, we used models with optimized hyper-parameters to report our final results for each dataset. As can be expected and without reporting all the hyper-parameters for every task, we observed that these parameters were significantly different across datasets.
	
	\begin{figure*}
	    \centering
	    \includegraphics[width=\textwidth]{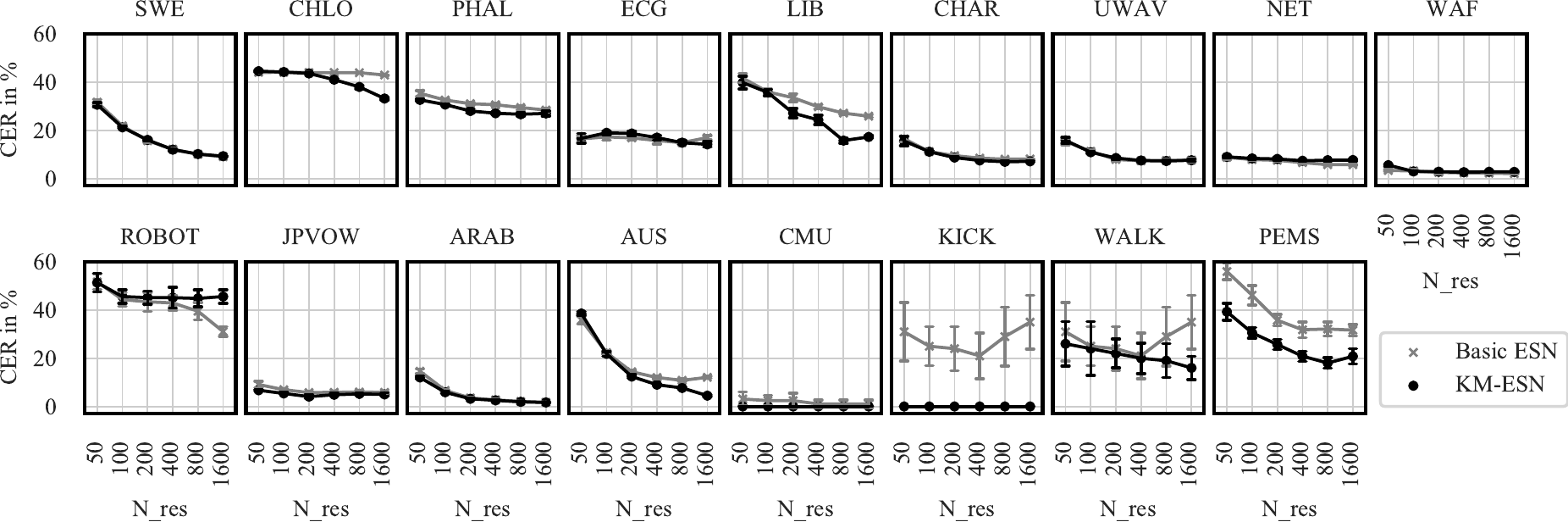}
	    \caption{Multi-dataset evaluation: Mean and standard deviation of the $\mathrm{CER}$ of the basic ESN and of the KM-ESN. For each dataset, the models were ten times randomly initialized, trained and evaluated. The KM-ESN outperformed the basic ESN in many datasets, both the basic ESN and the KM-ESN performed equivalently in various datasets and only in a few cases, the basic ESN performed better than the KM-ESN.}
	    \label{fig:MultiDatasetLargerReservoir}
	\end{figure*}

	\subsection{Results}
	\label{subsec:Experiment2Results}
    Fig.\ \ref{fig:MultiDatasetLargerReservoir} shows the mean value and standard deviation of $\mathrm{CER}$ for the different datasets for different reservoir sizes and \num{10} instances of each model. In order to investigate the impact of random initializations on the performance we repeated the training procedure ten times. In each run, all weights of the basic ESN and the reservoir weights of the KM-ESN were randomly initialized, trained using the training datasets and finally evaluated on the test sets. Fig.\ \ref{fig:MultiDatasetLargerReservoir} shows that the KM-ESN outperformed the basic ESN in many datasets, that both the basic ESN and the KM-ESN performed equivalently in various datasets and only in a few cases, the basic ESN performed better than the KM-ESN. Since for some datasets and models, the standard deviation of the loss was so low, the error bars are not always visible. However, overall, the standard deviation of the loss of the KM-ESN is often lower than that of the basic ESN. Thus, as in the previous experiment, the KM-ESN is more robust against random initalizations.

    The KM-ESN outperformed the basic ESN in particular for very high-dimensional datasets, such as PEMS, KICK and WALK, which all have more than \num{60} features. Another case, in which the KM-ESN performed remarkably better than the basic ESN is the two-dimensional LIB dataset, where the features are uniformly distributed in a first glance. However, the features can still be well clustered and thus, clustering the features still improves the performance of the KM-ESN compared to the basic ESN in case of larger reservoirs. In the ARAB dataset, which consists of extracted MFCC features from Arabian spoken digits, the smaller KM-ESNs slightly outperformed the basic ESNs, whereas for larger reservoirs, the performance became more and more similar. Since this is a dataset, in which features can be linearly separated, we postulate that the KM-ESN works particularly well for linearly separable datasets. Interestingly, in case of the datasets JPVOW and AUS, we can see that the KM-ESN not only outperformed the basic ESN, but it also allows to further increase the reservoir size without risk of overfitting.

	In case of the CHLO and the PHAL datasets, the KM-ESN slightly outperformed the basic ESN. Since these datasets have one-dimensional input features, this means that, given linearly separable features, the $K$-Means algorithm can cluster them. In fact, both CHLO and PHAL show patterns that make it possible to determine at least a few clusters, whereas the one-dimensional features of the SWE dataset are more normal distributed. Thus, the resulting centroids of the $K$-Means algorithm are concentrated at the maximum of the distribution.
    
    In case of the ROBOT dataset, the performance of the KM-ESN was relatively low compared to the basic ESN. According to Table~\ref{tab:MultiDatasets}, ROBOT was the smallest dataset to be considered in this study. It consists of less than \num{1500} samples. That means that, for the largest reservoir with \num{1600} neurons, we were \textit{forced} to introduce sparsity to the KM-ESN. Furthermore, $K$-Means algorithms are known to be sensitive against outliers. Applying Principal Component analysis (PCA) on the ROBOT dataset showed a large amount of outliers in this dataset. This, in turn, means that it is likely that the $K$-Means algorithm found a lot of centroids in a very small part of the feature space and thus many neurons in the KM-ESN received a lot of input information in an almost equivalent way. These could be the reasons why the KM-ESN performed worse than the basic ESN on the ROBOT dataset.
	
	\section{Conclusions and Outlook}
	\label{sec:ConclusionsAndOutlook}
	
	We presented an effective way to initialize the input weights of Echo State Networks using the unsupervised $K$-Means algorithm. Motivated by the fact that passing feature vectors to a reservoir neuron is closely related to the cosine similarity, we used the centroids of the $K$-Means algorithm as input weights. This controls the neuron activation such that they were high only if the feature vector and a centroid were similar. We showed that the input weight distribution of the basic ESN and the KM-ESN is significantly different, because the weights of the latter one were trained using the $K$-Means algorithm and are thus basically the average value of subsets of the training samples. This supported the hypothesis that the $K$-Means algorithm indeed supplied the KM-ESN with beneficial information about the dataset.
	
	We demonstrated that the KM-ESN model outperformed basic ESNs on various datasets. First of all, we studied the impact of replacing randomly initialized input weights with centroids obtained by the $K$-Means algorithm. Based on our experiments, the input weights are no more uniformly distributed between $\pm 1$ and the initialization is driven by the data. Also the $K$-Means-based ESN was more robust compared to random initialization, and the performance of the KM-ESN was higher than the performance of a basic ESN in particular for small numbers of neurons. In case of the multi-dataset evaluation, we have presented different use-cases together with suggestions when to prefer the KM-ESN or a basic ESN. It turned out that the KM-ESN is in particular useful for very high-dimensional datasets with a sufficient number of samples to train the $K$-Means algorithm. However, if the features have outliers, such as in the ROBOT task, the KM-ESN was less successful than the basic ESN, which by itself did not perform well. 

	As the $K$-Means-based input weight initialization is both data-driven and unsupervised, we obtain a set of input weights that is optimized for a given dataset. In general, if we would use a basic ESN for two different datasets with different features (e.g.\ audio features and sensor data) but with the same $N^{\mathrm{in}}$, we would simply use the same random set of input weights. However, since the features have completely different characteristics, it is likely that task-dependently adapted input weights boost the performance of ESNs. Such an adaptation is the main reason to support KM-ESN. Furthermore, we can easily extend a given dataset with more data, e.g.\ video recordings with different camera positions or with additional objects and kinds of noise without needing labels to help the KM-ESN to generalize towards unknown situations.
	
	In the future, one would analyze the capability of the proposed technique to solve more complex tasks, such as phoneme recognition in spoken language or multipitch tracking in music signals. It would also be interesting to determine the capability of predicting time-series. Another follow-up work would be to investigate ways for pre-training the reservoir weights as well. 
	Furthermore, it would be interesting to study whether the proposed KM-ESN is a universal approximator for dynamic systems according to \cite{src:Wang-17b}.
	
	Code examples for the two experiments are publicly available in our Github repository (\href{https://github.com/TUD-STKS/PyRCN/}{https://github.com/TUD-STKS/PyRCN/}). 

    \bibliographystyle{IEEEtran}
	\bibliography{refs}
	
	\appendices
    \section{Hyperparameter optimization}
    \label{sec:SequentialVsRandomized}
    
    Since the hyperparameter optimization explained in Sec.\ \ref{subsec:Experiment1OptimizatiofTheESNs} and proposed in \cite{src:Steiner-20a} was originally introduced for the basic ESN and successfully applied to various speech, music and image recognition tasks, we compared whether it can be used for the KM-ESN as well. 

    To do so, we compared two optimization strategies on the video classification task (Sec.\ \ref{sec:Experiment1DoorStateRecognition}): 

    \begin{enumerate}
        \item Sequential optimization as described in Sec.\ \ref{subsec:Experiment1OptimizatiofTheESNs} using the parameters in Table \ref{tab:OptimizationSearchSpace}. Only $\alpha_{\mathrm{bi}}$ was optimized with a 1D grid search. All other steps were randomized searches. We needed \num{321} optimization steps in total.
        \item Fully randomized optimization by jointly exploring the search space consisting of all parameters in Table \ref{tab:OptimizationSearchSpace}. We evaluated \num{2000} parameter combinations in total.
    \end{enumerate}

    Since we used 5-fold cross validation, each parameter combination (regardless sequential or joint optimization) was evaluated five times.
    
    \begin{table}[!htb]
    	\renewcommand{\arraystretch}{1.3}
        \caption{Search spaces for the sequential optimization based on the steps proposed in \cite{src:Steiner-20a}, and for the joint randomized search.}
        \label{tab:OptimizationSearchSpace}
        \centering
        \begin{tabular}{llll}
        \toprule
        Parameter & \multicolumn{2}{c}{Search space} & Iterations \\
		\midrule
		$\alpha_{\mathrm{u}}$ & \numrange{e-3}{1} & uniform & \multirow{2}{*}{\num{200}} \\
		$\rho$ & \numrange{0}{2} & uniform &  \\
		\midrule
		$\lambda$ & \numrange{e-5}{1} & loguniform & \num{50} \\
		\midrule
		$\alpha_{\mathrm{bi}}$ & \numrange{0}{2} & uniform & \num{21} \\
		\midrule
		$\epsilon$ & \numrange{e-5}{10} & loguniform & \num{50} \\
		\bottomrule
        \end{tabular}
    \end{table}
    
    From the results in Table \ref{tab:FinalHyperParams} can be seen that the sequential optimization and the joint randomized search in case of the basic ESN led to similar hyper-parameters. There are strong differences between the spectral radii and the bias scalings. The final loss is comparable as well. 
    \begin{table*}[!htb]
    	\renewcommand{\arraystretch}{1.3}
    	\caption{Optimized hyper-parameters and final loss values for the basic ESN and for the dense KM-ESN ($N_{res}<\num{200}$) after the joint randomized search and the sequential optimization.}
    	\label{tab:FinalHyperParams}
    	\centering
    	\begin{tabular}{lllll}
    		\toprule
    		& \multicolumn{2}{c}{Basic ESN} & \multicolumn{2}{c}{KM-ESN} \\
    		Parameter & random & sequential & random & sequential \\
    		\midrule
    		Input scaling $\alpha_{\mathrm{u}}$ & \num{0.95} & \num{0.85} & \num{0.03} & \num{0.06}  \\
    		Spectral radius $\rho$ & \num{0.14} & \num{0.05} & \num{1.05} & \num{0.08}  \\
    		Leakage $\lambda$ & \num{0.03} & \num{0.04} & \num{0.14} & \num{0.08}  \\
    		Bias scaling $\alpha_{\mathrm{bi}}$ & \num{0.21} & \num{0} & \num{0.13} & \num{0}  \\
    		\midrule
    		Regularization $\epsilon$ & \num{15e-4} & \num{13e-4} & \num{11e-4} & \num{6e-4}  \\
    		\midrule
    		Final loss & \num{7.2e-2} & \num{7.0e-2} & \num{6.8e-2} & \num{4.8e-2}  \\
    		\bottomrule
    	\end{tabular}
    \end{table*}

    In case of the KM-ESN, we can see that the spectral radii are particularly different. In case of the sequential optimization, the spectral radius is close to zero, whereas the spectral radius in case of the joint randomized search is close to one. However, the losses differ more than in case of the basic ESN. Table \ref{tab:FinalHyperParams} shows that it decreased much more in case of the sequential search.

\end{document}

%% file: figs/results/video_dataset/kmeans_opt_video.tex
\begin{tikzpicture}
\begin{axis}[
width  = 0.46*\textwidth,
height = .2\textwidth,
cycle list name=color,
major x tick style = transparent,
ymajorgrids = true,
xmajorgrids = true,
ylabel = {$\mathrm{SSE}$},
xlabel={$K$},
xmode=log,
log ticks with fixed point,
scaled x ticks = false,
scaled y ticks = false,
xmin=2,
xmax=2000,
ymin=10000,
ymax=4200000,
label style={font=\small},
ticklabel style = {font=\small},
legend cell align=left,
legend columns=2,
legend style={
	at={(0.5,1.0)},
	anchor=south,
	column sep=1ex,
	font=\scriptsize,
},
]
\addplot[black, very thick, no markers]
table {%
2 4129520.1150603155
3 2675987.214773209
4 2129150.2797099296
5 1787096.0806590584
6 1538901.9985508178
7 1350314.7858886719
8 1247260.838508869
9 1133134.0718756034
10 1085573.2503530146
20 688486.7278448677
30 502153.74121014285
40 435848.5512318774
50 377125.406600825
60 357359.87583535403
70 331299.2554761263
80 311905.0589809085
90 293554.4098665418
100 285807.8935592438
200 219894.91349089734
300 199323.8434175164
400 155538.04514839678
500 138850.4751758618
600 128880.32758227867
700 123432.36882255868
800 117717.33544241691
900 151496.5754852751
1000 143832.80225897723
2000 115281.28831628563
};
\addplot+[ycomb, white!60!black, thick, mark=o] plot table%
{
50 377125.406600825
100 285807.8935592438
200 219894.91349089734
400 155538.04514839678
800 117717.33544241691
};
\legend{
$\mathrm{SSE}$, Analyzed $K$s
}
\end{axis}
\end{tikzpicture}

%% file: figs/results/video_dataset/video_final_score.tex
\begin{tikzpicture}
\begin{axis}[
width  = 0.46*\textwidth,
height = .2\textwidth,
cycle list name=color,
major x tick style = transparent,
ymajorgrids = true,
xmajorgrids = true,
ylabel = {$\mathrm{FER}$ in \si{\percent}},
xlabel={$N_{\mathrm{res}}$},
xtick={50, 100, 200, 400, 800, 1600, 3200},
xmode=log,
log ticks with fixed point,
scaled x ticks = false,
scaled y ticks = false,
xmin=46,
xmax=1700,
ymin=0.1,
ymax=13,
legend cell align=left,
legend columns=3,
legend style={
	at={(0.5,1.0)},
	anchor=south,
	column sep=1ex,
	font=\scriptsize,
},
label style={font=\small},
ticklabel style = {font=\small},
]
\addplot [ultra thick, white!80!black, mark=*, mark size=2pt] 
  plot [error bars/.cd, y dir=both, y explicit, , error bar style={ultra thick}, error mark options={ultra thick, rotate=90}]
  table [y error plus expr=\thisrow{y_max} - \thisrow{y},
         y error minus expr=\thisrow{y} - \thisrow{y_min},] {
x y y_min y_max
50 6.2030857 0.8149702 12.8597828
100 0.847237 0.5748343 1.7040505
200 2.6362976 0.5794061 4.9545474
400 4.8156632 4.7921309 4.8513227
800 4.7515869 4.7252393 4.7904465
1600 4.8122705 4.6987714 5.4610465
};
\addplot[ultra thick, white!50!black, mark=+, mark size=2pt]
  plot [error bars/.cd, y dir=both, y explicit, , error bar style={ultra thick}, error mark options={ultra thick, rotate=90}]
  table [y error plus expr=\thisrow{y_max} - \thisrow{y},
         y error minus expr=\thisrow{y} - \thisrow{y_min},] {
x y y_min y_max
50 4.1938604 0.6256046 7.8311253
100 0.4958397 0.4364795 0.5406667
200 0.444468 0.4215612 0.4814749
400 3.6623853 0.6667501 4.8664816
800 4.7870538 4.7644599 4.7221113
1600 4.7472076 4.8005525 4.7728815
};
\addplot[ultra thick, black, mark=x, mark size=2pt]
  plot [error bars/.cd, y dir=both, y explicit, , error bar style={ultra thick}, error mark options={ultra thick, rotate=90}]
  table [y error plus expr=\thisrow{y_max} - \thisrow{y},
         y error minus expr=\thisrow{y} - \thisrow{y_min},] {
x y y_min y_max
50 4.1938604 0.6256046 7.8311253
100 0.4958397 0.4364795 0.5406667
200 0.444468 0.4215612 0.4814749
400 0.5890307 0.4930245 0.7608314
800 0.6067642 0.5014461 0.7324386
1600 0.6366729 0.4831592 0.7800808
};
\legend{
Basic ESN, KM-ESN (d.), KM-ESN (sp.),
}
\end{axis}
\end{tikzpicture}%